\documentclass[]{llncs}
\usepackage[T1]{fontenc}
\usepackage{graphicx}
\usepackage{subcaption}
\usepackage{booktabs}
\usepackage{amsmath}
\usepackage[title]{appendix}
\usepackage{amssymb}
\usepackage{tikz}
\usepackage{float}
\usepackage{listings}
\usepackage{url}
\usepackage{natbib}
\usepackage{adjustbox}
\usepackage{verbatim}

\usepackage{hyperref}

\begin{document}
%
\title{Limitations of Scalarisation in MORL: A Comparative Study in Discrete Environments}
%
%
%

\titlerunning{Limitations of Scalarisation in MORL}

\author{Muhammad Sa'ood Shah\inst{1}\orcidID{0009-0009-2452-7847} \and
Asad Jeewa\inst{1,2}\orcidID{/0000-0003-4329-8137}}

%
\authorrunning{M. Shah and A. Jeewa}

%

\institute{School of Mathematics, Statistics and Computer Science,\\ University of KwaZulu-Natal, Durban, South Africa \and Centre for Artificial Intelligence Research (CAIR), Durban, South Africa\\ \email{221021204@stu.ukzn.ac.za, jeewaa1@ukzn.ac.za}}

\maketitle

\begin{abstract}
Scalarisation functions are widely employed in MORL algorithms to enable intelligent decision-making. However, these functions often struggle to approximate the Pareto front accurately, rendering them unideal in complex, uncertain environments. This study examines selected Multi-Objective Reinforcement Learning (MORL) algorithms across MORL environments with discrete action and observation spaces. We aim to investigate further the limitations associated with scalarisation approaches for decision-making in multi-objective settings. Specifically, we use an outer-loop multi-policy methodology to assess the performance of a seminal single-policy MORL algorithm, MO Q-Learning implemented with linear scalarisation and Chebyshev scalarisation functions. In addition, we explore a pioneering inner-loop multi-policy algorithm, Pareto Q-Learning, which offers a more robust alternative. Our findings reveal that the performance of the scalarisation functions is highly dependent on the environment and the shape of the Pareto front. These functions often fail to retain the solutions uncovered during learning and favour finding solutions in certain regions of the solution space. Moreover, finding the appropriate weight configurations to sample the entire Pareto front is complex, limiting their applicability in uncertain settings. In contrast, inner-loop multi-policy algorithms may provide a more sustainable and generalizable approach and potentially facilitate intelligent decision-making in dynamic and uncertain environments.
\end{abstract}

\keywords{Multi-Objective Reinforcement Learning \and Scalarisation \and outer-loop multi-policy\and inner-loop multi-policy \and Discrete Environments}


\section{Introduction}
\label{intro}

Reinforcement learning has had tremendous success with single-objective problems \citep{DBLP:journals/corr/MnihKSGAWR13,lillicrap2015continuous,haarnoja2018soft}. However, in the real world, scenarios generally have concurrent objectives \citep{coello2007evolutionary,hernandez2012emergent}. These problems can be studied through multi-objective reinforcement learning (MORL), yet the MORL domain has been relatively underexplored \citep{hayes2022practical,felten2022metaheuristics}. The challenge lies in learning solutions that need to learn to balance trade-offs between the objectives, which are often conflicting.

Single-policy and multi-policy MORL algorithms are used to tackle multi-objective problems \cite{yang2019generalized}. Single policy methods aim to find an optimal solution that exhibits a specific trade-off between the objectives. However, numerous optimal solutions may exist for a given multi-objective problem, and user preferences will vary over time. Multi-policy algorithms were designed to learn many optimal solutions and provide users with various options.

There are two main types of multi-policy algorithms \citep{hayes2022practical,roijers2013survey,felten2022metaheuristics}. Outer loop approaches run a single policy algorithm multiple times with varying parameters to obtain diverse solutions. Inner-loop approaches are more dynamic and redefine the learning algorithm, facilitating the learning of multiple optimal solutions simultaneously.

Scalarisation functions (utility functions) are frequently employed in MORL algorithms to choose actions and have been widely adopted in outer-loop multi-policy methods \citep{felten2022metaheuristics,hayes2022practical}. The scalarisation function transforms the multi-dimensional rewards into a single scalar value, allowing the treatment of the task as a single-objective reinforcement learning problem \cite{10.1007/s10994-010-5232-5}. A limitation of this approach is that these functions often require weighting tuples, where objectives are assigned values, with objectives of greater importance having a higher weight. However, this introduces bias as this heavily influences the solution found, and 
the difficulty arises in selecting weights to sample the Pareto front accurately \cite{das1997closer}. Newer inner-loop multi-policy algorithms have been opposing the scalarisation approach and developing alternate innovative and dynamic alternatives \cite{hayes2022practical}. 

As highlighted by Vamplew et al., there is a lack of comparative studies of current MORL algorithms in the domain \cite{10.1007/s10994-010-5232-5}. Most research papers focus on developing new methods to solve previous methodological challenges. Furthermore, it was difficult to directly compare the performance of algorithms because there was no standardised API for the widely accepted implementation of MORL algorithms and environments until 2023 \citep{hayes2022practical,cassimon2022survey}. The introduction of MORL Baselines, a standardised API for reliably implemented MORL algorithms, and MO-Gymnasium, a standardised API for reliably implemented MORL environments, addresses these issues and has made it possible for a rigorous and consistent evaluation of MORL algorithms \cite{morlbaselines}. They called for future work in MORL to use this framework to conduct experiments and perform exhaustive hyperparameter tuning on the algorithms.


This study primarily analyses the limitations of outer-loop multi-policy methods, which often deploy scalarisation functions for decision-making. We aim to illustrate why scalarisation functions are less effective, particularly in scenarios where the shape of the true Pareto front is unknown and why inner-loop methods offer a more sustainable and reliable approach. 

The code for our experiments can be accessed from \href{https://anonymous.4open.science/r/morl-baselines-C815}{this repository} under "my experiments folder"

\section{Preliminaries}
\label{prelim}
Reinforcement learning (RL) is a machine learning technique in which we train agents by allowing them to interact with an environment. The agent is an entity that learns and makes decisions, while the environment is the world with which our agent interacts. We reward the agent whenever it performs desirable actions and punish it for undesirable ones. This leads to a sequence of actions that lead to success or failure, from which our agent learns. RL has been widely used and incorporated in robotics, image processing, recommendation systems, gaming, traffic control management, and healthcare \cite{Leavey}. 

An action space $A$ is all the possible actions an agent can choose from in an environment, while an observation space (or state space $S$ is all the possible states or observations that the agent can perceive from the environment. A discrete action space means that the agent has a finite set of actions from which to choose. A discrete observation space is a finite set of distinguishable states that the agent observes \cite{Sutton1998}. RL agents learn to make decisions by optimising behaviour within a Markov Decision Process (MDP), defined as $M=(S,A,P,R,\gamma)$, where  P is the transition function (not used by model0free algorithms used in this work, $R$ is a scalar reward function, and 
$\gamma$ is a discount factor for future rewards. A Multi-Objective MDP (MOMDP) extends this by using a vector-valued reward $R:S \times A→\mathbb{R}^n$ to capture multiple, potentially conflicting objectives.

\textbf{Multi-Objective Reinforcement Learning (MORL):}
While RL aims to maximise the reward return where the scenario only has a single objective, MORL has multiple objectives. This means that the reward function in MORL is no longer a scalar value but rather a vector \cite{10.1007/s10994-010-5232-5}. In RL, the goal is to learn a single optimal policy; however, with MORL, there could be multiple optimal policies. A policy or solution is considered optimal or non-dominated if it cannot improve one objective without worsening another \cite{10.1007/s10994-010-5232-5}. The Pareto front represents the set of optimal trade-offs between objectives. Each optimal solution forms a point that approximates the Pareto front.
\begin{figure}[h]
    \centering
    \includegraphics[width=0.5\linewidth]{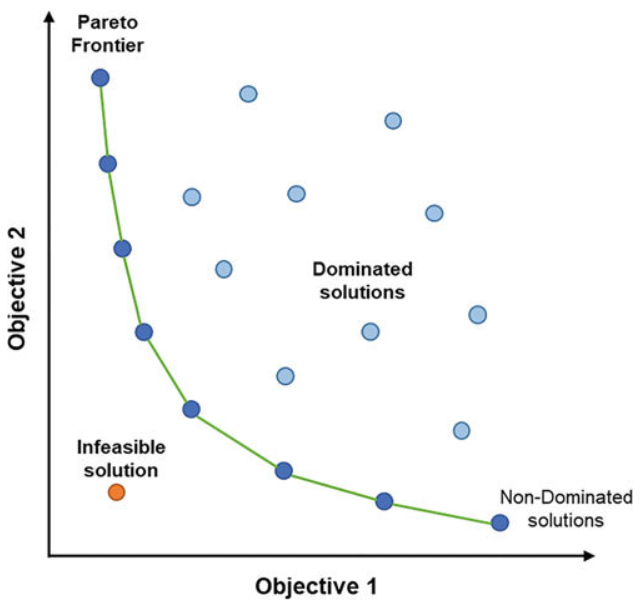}
    \caption{Pareto front in multi-objective learning}
    \label{PF example}
\end{figure}

In \autoref{PF example}, if a line segment connects two points on a curve and any part of the segment lies outside the curve, those points are considered nonconvex.

\textbf{MO Q-Learning function} 

MO Q-Learning function is a single policy MORL algorithm proposed by van Moeffart et al. \cite{moq}. For each state-action pair $(s,a)$, we have a vector of q values corresponding to each objective.
\begin{equation}
\hat{Q}(s, a) = [Q_1(s, a), Q_2(s, a), \dots, Q_n(s, a)]
\end{equation}
where there are n objectives and, $\hat{Q}_n(s,a)$ is the Q-value for the $n^{th}$ objectives

In particular, we study the MO Q algorithm with the following scalarisation functions:

\textbf{Linear Scalarisation:} 
 We assign weights ranging from [0,1] for each objective, where the sum of the total weights must be added to 1 \cite{moq}. We apply a weighting sum mechanism to obtain our scalarised Q values SQ(s, a) for each state-action pair (s, a).
\begin{equation}
SQ(s, a) = \sum_{o=1}^{m} w_o \cdot \hat{Q}(s, a, o)
\end{equation}
An underlying critique of this function is the inability to learn non-convex solutions of the Pareto front \cite{moq}. We chose to utilise this function to validate its performance in the existing literature. Moreover, it was used to further assess the strengths and limitations of scalarisation functions. 

\textbf{Chebyshev Scalarisation:}
In addition to assigning weights to each objective, the Chebyshev scalarisation utilises a utopian point, $z_o$, as a reference point when selecting actions. The utopian point stores the best value for each objective o and a small constant, $\tau$ i.e. \begin{equation}
z_o = f_{\text{best}}(o) + \tau
\end{equation}
The scalarised Q value is calculated as follows:
\begin{equation}
SQ(s, a) = \max_{o=1,\dots,m} w_o \cdot |\hat{Q}(s, a, o) - z_o|
\end{equation}
The action that is considered best is the minimal SQ -value
\begin{equation}
\text{greedy}(a') = \min_{a'} SQ(s, a')
\end{equation}

Although it is not guaranteed to converge to a Pareto optimal solution \cite{perny2010finding,roijers2013survey}, it has shown its efficacy in limited MORL experiments \cite{moq,van2013hypervolume}. We employed this scalarisation function with the MO Q-Learning algorithm and further investigated its performance.

\textbf{Pareto Q-Learning}
\label{pql}
PQL is an approach by which the algorithm learns multiple optimal policies simultaneously. It works by separating Q values from immediate and future discounted rewards, helping to keep track of short-term gains and long-term goals. Opposing MORL algorithms, which commonly employ scalarisation functions for action selection, it uses an inbuilt decision mechanism that utilises multi-objective axiomatic measures such as hypervolume measure, cardinality indicator and Pareto dominance relation to select actions. 
PQL is a tabular algorithm that stores Q values for each state-action pair. This approach makes it memory-intensive and results in scalability issues in complex environments with larger state spaces, such as Four-Room.

\section{Literature Review}
\label{lit}

\subsection{Challenges with MORL}

There are many existing challenges in MORL. In particular, this domain has been relatively underexplored, with a lack of standardised APIs available to experiment with past algorithms and environments and limited comparative studies of current work. 

Although most real-world tasks involve multiple objectives, most RL research has focused on single-objective RL \cite{felten2022metaheuristics}. Although these algorithms have succeeded in RL, they often combine objectives into a scalar value when scaled to scenarios with multiple objectives, which does not directly address the multi-objective problem \cite{hayes2022practical}. This is concerning, as real-world problems are inherently multiple-objective in nature. 

It is common for authors not to release their code publicly, while published code is often poorly maintained.\citep{abels2019dynamic,yang2019generalized,xu2020prediction}. This makes it challenging to evaluate algorithms. Moreover, Hayes et al. pointed out the lack of standardised APIs to access existing MORL environments \cite{hayes2022practical}. With the lack of standardisation, direct comparisons of MORL algorithms are scarce as differences may stem from inconsistent implementation rather than algorithm performance \cite{vamplew2017morl}. This has led to a scarcity of comparative analysis in the MORL literature, which was initially highlighted by Vamplew et al. in 2011 \cite{10.1007/s10994-010-5232-5}. 

In RL, there have been successful initiatives to address these issues, such as OpenAI Gym and Stable-Baselines 3 \citep{brockman2016openai, raffin2021stable}. In MORL, an initial attempt to address standardisation was the Glue Benchmark suite \cite{vamplew2017morl}. However, it proved impractical as it was only implemented in Java.

MORL-Baselines and MO-Gynasium are standardised libraries for MORL environments and algorithms introduced by Felten et al. \cite{morlbaselines}. All algorithms in MORL-Baselines are regularly maintained and compatible with the environments of MO-Gymnasium. The authors called for future research to utilise their framework to gain insight into MORL algorithms. My research aims to use these frameworks to evaluate selected MORL algorithms across discrete-spaced environments.

\subsection{MORL Algorithms}

Numerous algorithms have been developed in this domain. In this subsection, we study some of the MORL algorithms implemented. 

Van Moffaert et al. introduced a single-policy algorithm called MO Q-Learning that could employ either a linear or a non-linear scalarisation function \cite{moq}. They highlighted issues with the linear scalarisation function and demonstrated how a non-linear scalarisation function, the Chebyshev function, could find solutions in non-convex areas of the Pareto front. 

Van Moffaert, K. and Nowe, A. highlighted the drawbacks of single-policy methods and presented an inner-loop multi-policy algorithm, Pareto Q-Learning. This algorithm does not require weighting tuples and can converge with the entire Pareto optimal set \cite{pql}. Furthermore, the authors exhibited an epsilon-decaying exploration strategy that begins with a high epsilon value, which decays as training progresses, allowing the agent to explore the environment initially and then exploit the best actions. The results of Pareto Q-Learning proved its efficacy by converging to the Pareto fronts of the testing environments. We employed this algorithm in the Deep Sea environments as a baseline inner-loop multi-policy approach to illustrate its superiority over outer-loop multi-policy methods. Additionally, all algorithms in the experiments used the epsilon-decaying exploration approach. Our paper investigates why focus should be placed on inner-loop multi-policy methods rather than outer-loop policy methods. 

Many of the newer MORL algorithms are being shifted towards inner-loop multi-policy approaches. Basaklar et al. proposed a multi-policy method that does not require prior information about user preference and can be extended to continuously spaced environments, which was not achievable with previous multi-policy algorithms \cite{basaklar2022pd}. Reymond et al. implemented a multi-policy algorithm, the Pareto Condition Network, which utilises a single neural network to find optimal solutions \cite{reymond2022pareto}. Other notable algorithms are Multi-Objective Reinforcement Learning Based on Decomposition MORL/D \cite{felten2024multi}, Optimistic Linear Support (OLS) \cite{roijers2016multi} and CAPQL \cite{lu2023multi}. These algorithms can be explored in future work.

The influx of new algorithms shows the expansion of the MORL community with new innovative, robust, and scalable algorithms. This paper focuses on critically evaluating a single policy algorithm, MO Q-Learning, as an outer-loop method. We explore the limitations of this approach, as it is commonly used in MORL \cite{felten2022metaheuristics}.
 
\subsection{MORL Environments}

MORL environments play an important role in benchmarking algorithms. Apart from the standardisation issues mentioned above, there are issues arising from the complexity levels of the environments \cite{hayes2022practical}. 

In MORL, environments are typically characterised by their action and observation spaces. Some environments have discrete action and observation spaces, meaning the agent can only choose actions from a finite set of options. Deep-Sea Treasure is one of the most widely popular discrete space environments \cite{hayes2022practical}. It was originally proposed by Vamplew et al. It was one of the first MORL environments introduced and, since then, has been widely used to benchmark algorithm \cite{10.1007/s10994-010-5232-5}. It has gained significant attention due to its small state and action space and because it only has two objectives \cite{cassimon2022survey}. This environment was used as a baseline for benchmarking the paper's selected MORL algorithms.

Another environment with discrete spaces is the Four-Room. This environment was originally used in the literature on Successor Features (SFs) \cite{barreto2017successor}. Alegre et al. adapted this environment into MORL to demonstrate that any problem in SFs can be represented in the MORL framework \cite{alegre2022optimistic}.In this domain, the agent tries to collect shapes with positive rewards and avoid those with negative rewards while trying to reach the goal state. The Four-Room environment has yet to be explored and features an unknown Pareto front, which is why this environment was studied in this paper. Many other environments exist with discrete action and observation spaces \cite{barreto2017successor,barrett2008learning,vamplew2021potential}.

Environments can also have continuous action and observation space, meaning that states or actions are represented by continuous values \cite{hung2022q,10.1007/s10994-010-5232-5}. However, the environments studied in this paper are restricted to discrete spaces.

In conclusion, the growth of the MORL domain has led to resolutions of some of the previous challenges. Regarding MORL algorithms, there is a shift towards inner-loop multi-policy approaches that overcome past limitations. An area of improvement in the field includes introducing more challenging environments that are representative of real-life tasks at some level and conducting a comparative analysis of algorithms.

\section{Methods}
\label{methods}

\subsection{MORL Algorithms}

The following algorithms were explored in this paper: MO Q-Learning with Linear Scalarisation and Chebyshev  Scalarisation~\cite{moq} as well as Pareto Q-Learning~\cite{pql}.

The hyperparameters of all algorithms were tuned. We highlight the most important hyperparameters in~\autoref{tab:hyperparams} and provide further details in the appendix.
\begin{table}[h]
    \centering
    \begin{tabular}{l c c c c}
        \toprule
        Environment & Timesteps & $\epsilon_{initial}$ & $\epsilon_{final}$ & Gamma \\
        \midrule
        DST Concave & 400,000 & 1.0 & 0.1 & 0.9 \\
        Four-Room & 800,000 & 1.0 & 0.1 & 0.99 \\
        \bottomrule
    \end{tabular}
    \caption{Hyperparameters used for different environments}
    \label{tab:hyperparams}
\end{table}

The $\tau$ parameter for MO Q-Learning with Chebyshev scalarisation was set to 4 in DST and 6 in Four-Room environments \footnote{The $\tau$ parameter tended to find exact solutions even with different values.}. 

\subsection{Testing Environments}
\label{env}
\begin{figure*}[htbp]
    \centering
    \begin{subfigure}[b]{0.4\textwidth} 
        \centering
        \includegraphics[width=0.8\textwidth]{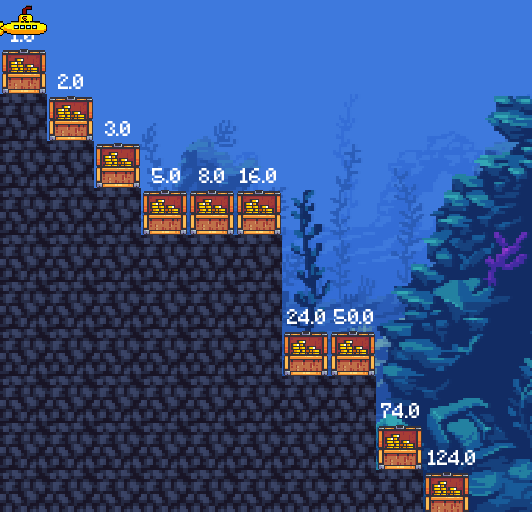}
        \caption{DST Concave}
        \label{fig:y_equals_x}
    \end{subfigure}
    \hspace{0.05\textwidth} 
    \begin{subfigure}[b]{0.4\textwidth} 
        \centering
        \includegraphics[width=0.8\textwidth]{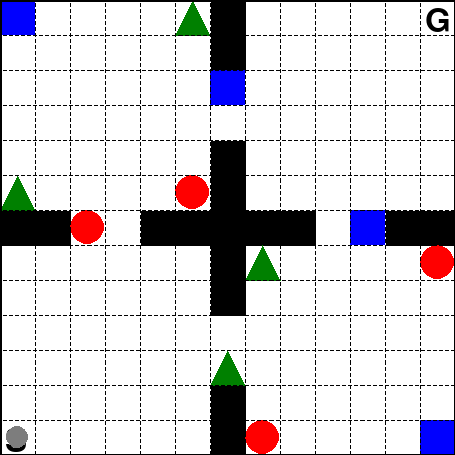}
        \caption{Four-Room}
        \label{fig:five_over_x}
    \end{subfigure}
    \caption{Testing environments used}
    \label{fig:multi_environment}
\end{figure*}
\textbf{Deep Sea Treasure Concave}

The agent controls a submarine with the aim of maximising the treasure value obtained while minimising the number of steps taken. The farther away the treasure, the more valuable it is. For each step the agent takes, it receives a -1 penalty \cite{10.1007/s10994-010-5232-5}. The shape of the Pareto front is concave, where each optimal path to each treasure is considered a non-dominated solution.

\textbf{Four-Room} 

In this domain, the agent tries to collect shapes with positive rewards while trying to reach the goal state \cite{barreto2017successor}. This environment has a much larger state-action space than the other environments mentioned in this paper. Furthermore, the true Pareto front for this environment is unknown.  This is because the balance between the objectives is not directly known, and there are many possible solutions that different exploration strategies can obtain. The four-room environment has not yet been studied in MORL and is an adaption of a Single-Objective RL environment, which was originally used for studying Successor Features \citealp[]{alegre2022optimistic}.

\subsection{Evaluation Metrics}

We used the following metrics to analyse the performance of algorithms that are widely used in the MORL literature:

\textbf{Hypervolume} The hypervolume metric is widely used in Multi-Objective Optimisation literature to analyse Pareto approximation \citep{zitzler1999multiobjective,hayes2022practical}. It is defined as the volume that the non-dominated solutions cover from a reference point. The reference point is usually specified for extreme values, such as the lower or upper limit of the objectives' reward space, for the objectives~\cite{felten2022metaheuristics}. The reference point used in DST Concave was (0,-50) while in Four-room was (-1,-1,-1) \footnote{All the reference points used to calculate the Hypervolume for the environment were procured from a discussion in the MORL-Baselines repository on GitHub.}.

\textbf{Sparsity:} Sparsity measures the spread between the optimal solutions on the Pareto front \cite{deb2002fast}. Lower sparsity indicates a uniform, denser solution set with a more compact Pareto front coverage. Higher sparsity indicates more diverse solutions, but also gaps in the Pareto front coverage.

\textbf{Cardinality:} Cardinality shows the number of non-dominated (optimal) solutions found by the algorithm. The Cardinality of the True Pareto front reveals the maximum optimal solutions for a given environment, and ideally, we want our algorithms to achieve the same number of optimal solutions.

\textbf{Inverted Generalised Distance (IGD):} IGD measures the distance of the optimal solutions from the approximated Pareto front to the solutions in the true Pareto front. A lower IGD value indicates closer coverage of the actual Pareto front. This metric can only be used when the Pareto front of the environment is known. Therefore, it can be employed In DST Concave and DST Mirrored but not in Four-Room.

\subsection{Experimental Design Methodology}

We conducted a parameter search to find the set of optimal hyperparameters for the MORL algorithm for each environment. More details can be found in the Appendix \ref{append}

An epsilon decaying exploration approach was employed in all experiments, starting around 1 and gradually decaying to 0.1 as training progressed. 

\subsubsection{MO Q-Learning}
\label{MO Q experimental design}

The approach to adapt the MO Q-Learning algorithm as an outer-loop multi-policy algorithm was based on the approach used by Van Moffaert et al. \cite{van2013hypervolume}.

The total number of objectives for each environment was inspected, and weight combinations were calculated using a step size of 0.1, ensuring that the weights were summed to 1. Each unique weight combination was considered a distinct configuration. Environments with two objectives yielded 11 unique configurations, while environments with 3 objectives increased to 64 configurations.

Each configuration was trained in a fixed number of timesteps, influenced by the complexity of the environment. Performance was tracked every 1000 timesteps by testing the learnt policy, obtaining the discounted vector reward return, and storing it in a set. This set is referred to as the Pareto front approximation set for that specified evaluation iteration, and it stores the discounted vector rewards at that iteration for all configurations. The convergence to the approximated Pareto front of these sets was examined by employing Hypervolume, Sparsity, Cardinality, and IGD metrics.

\subsubsection{Pareto Q-Learning}

The performance evaluation of PQL, an inner-loop multi-policy approach, was provided by the MORL-Baselines framework \cite{morlbaselines}. The algorithm's performance was evaluated every 1000 steps, and metrics such as Hypervolume, Sparsity, Cardinality, and IGD were utilised to analyse the convergence of the approximated Pareto fronts. 

All results were collected and averaged over 10 trials\footnote{We used different seed numbers (42,43,44,45,46,47,48,49,50,51) for each trial to produce reliable and reproducible results.}.

\section{Results and Discussion}
\label{experiments}
\subsection{Experiment 1: Deep-Sea Treasure Concave}\label{exp_1}

We used this environment as a benchmark to illustrate the challenges of using linear scalarisation in finding solutions in non-convex areas of the front and assess the capabilities of the Chebyshev scalarisation function. The results provided valuable information and served as a guide for the remaining experiments.

\begin{figure}
    \centering
    \begin{subfigure}[t]{0.48\textwidth}
        \centering
        \includegraphics[width=1\linewidth]{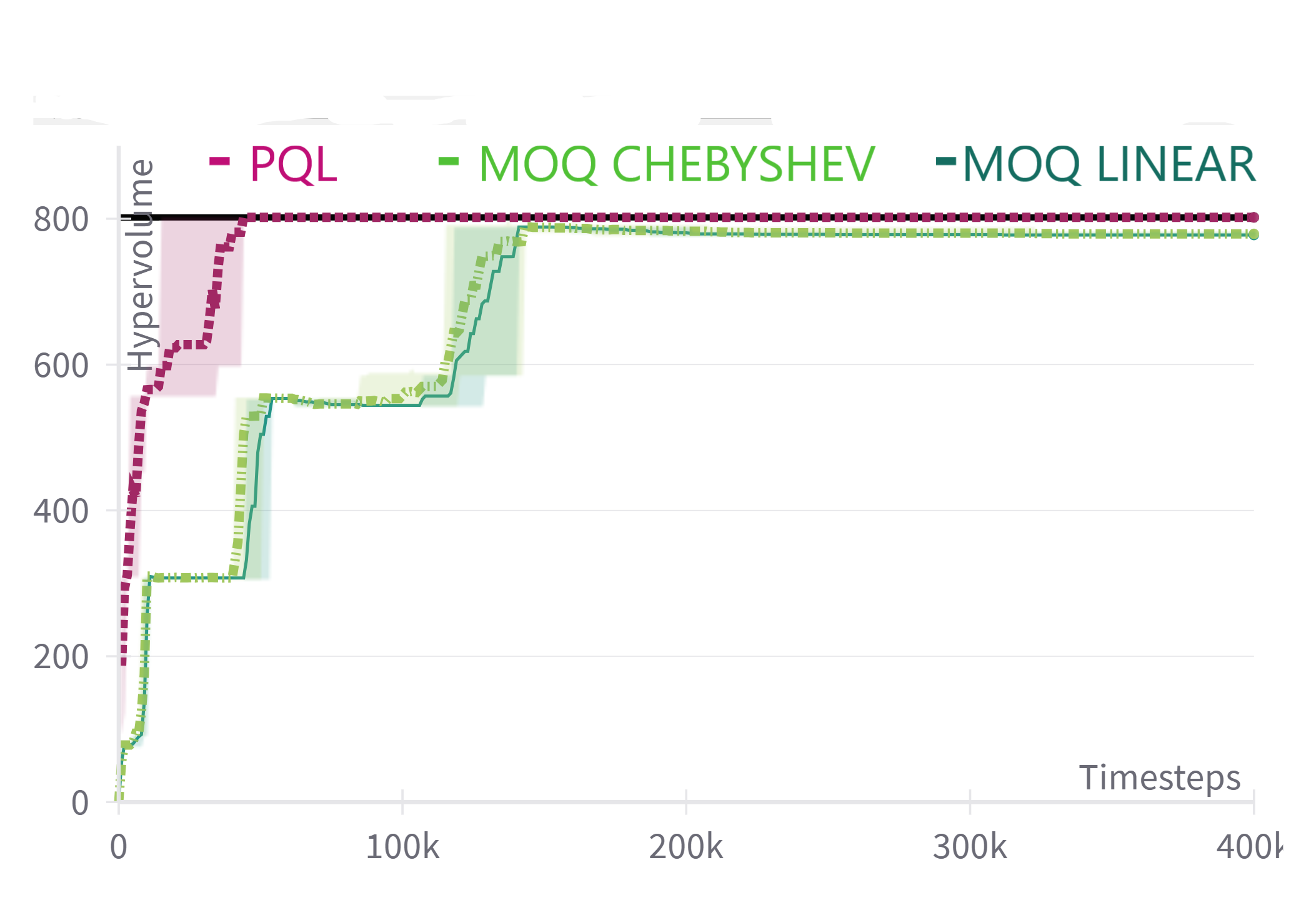}
        \caption{Hypervolume}
        \label{dst_hyp}
    \end{subfigure}
    \hfill
    \begin{subfigure}[t]{0.48\textwidth}
        \centering
        \includegraphics[width=1\linewidth]{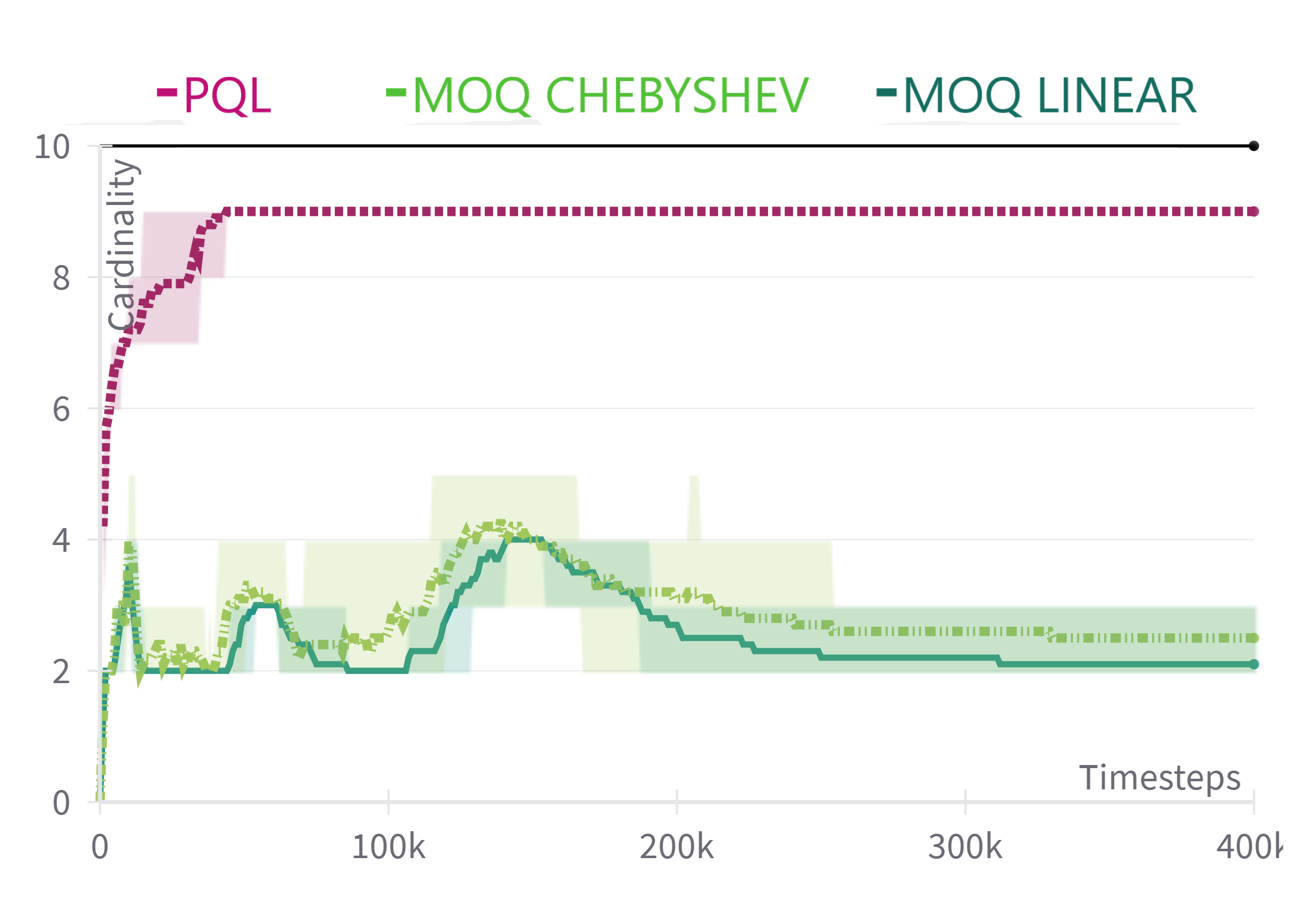}
        \caption{Cardinality}
        \label{dst_card}
    \end{subfigure}
        \begin{subfigure}[b]{0.45\linewidth}
        \centering
        \includegraphics[width=\linewidth]{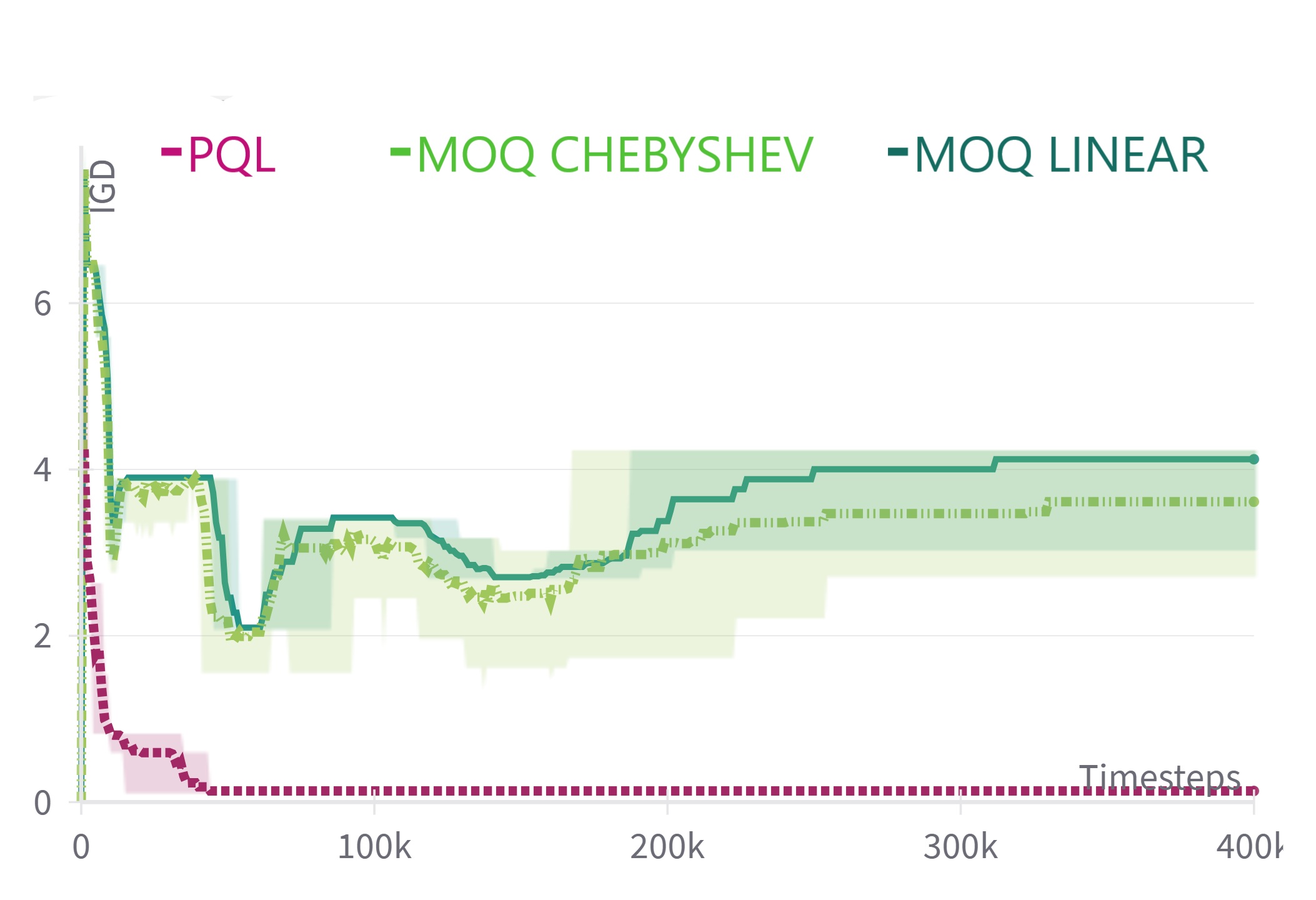}
        \caption{IGD}
        \label{dst_igd}
       
    \end{subfigure}
    \hfill
    \begin{subfigure}[b]{0.45\linewidth}
        \centering
        \includegraphics[width=\linewidth]{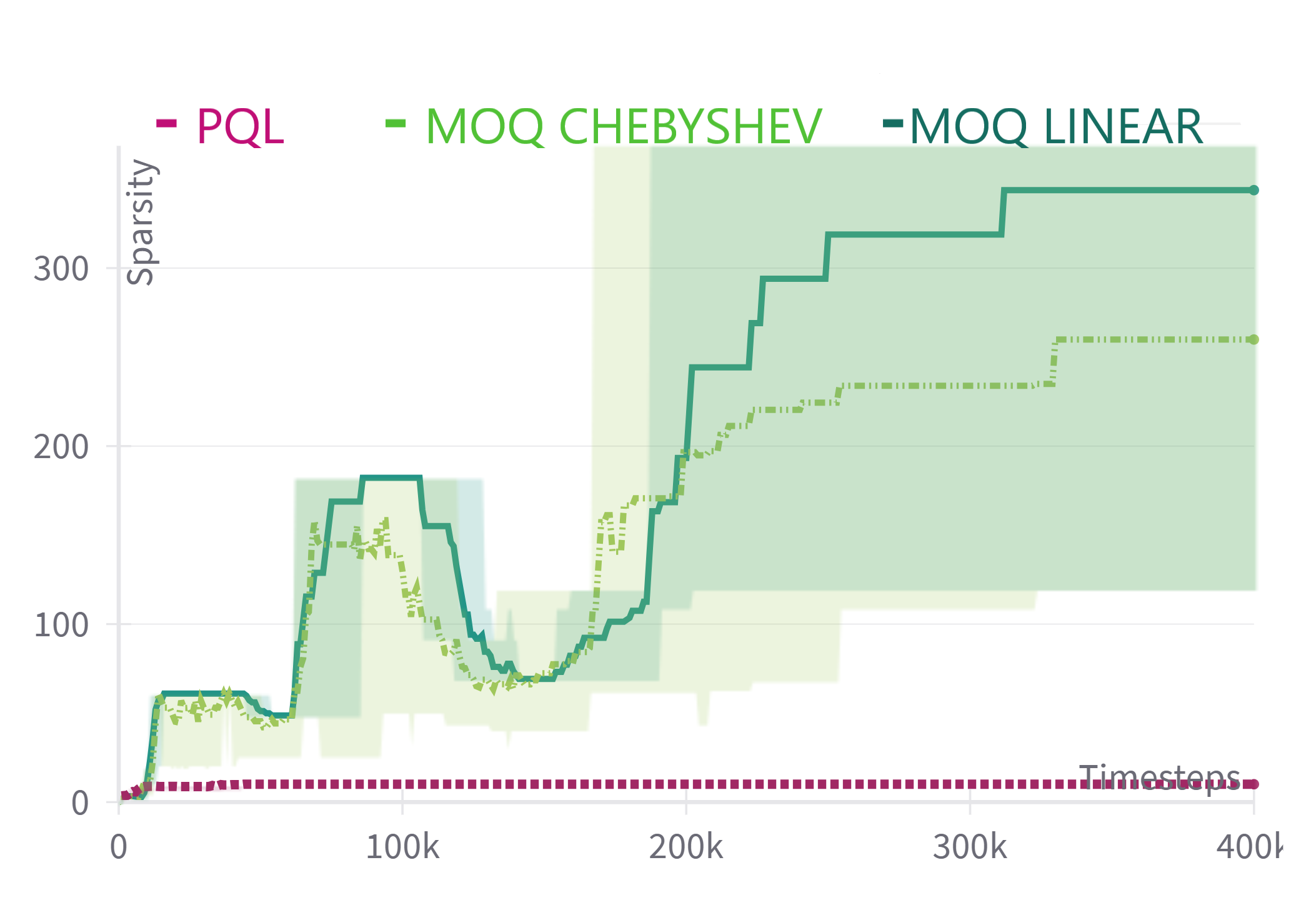}
        \caption{Sparsity}
        \label{dst_sparsity}
    \end{subfigure}
    \vskip\baselineskip
    \begin{subfigure}[b]{0.7\linewidth}
        \centering
         \includegraphics[width=0.8\textwidth]{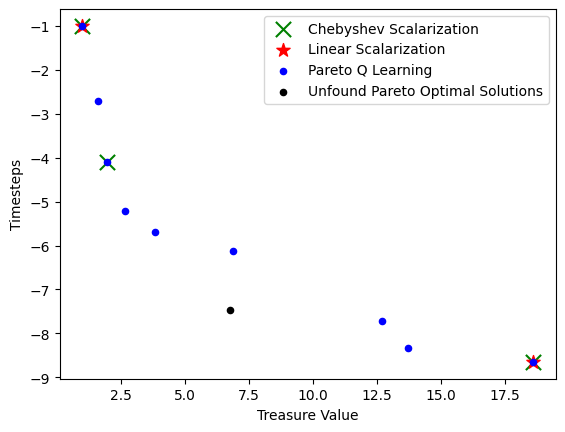}
        \caption{Pareto front approximations}
        \label{dst_pf}
    \end{subfigure}
    \caption{Performance metrics of algorithms in DST Concave}
    \label{dst_combined}    
\end{figure}

\autoref{dst_hyp} indicated that all the algorithms converged with similar hypervolume scores. However, PQL learnt much quicker and converged to its approximated Pareto front after 60,000 timesteps, while The linear and Chebyshev continued learning and only stagnated after 150,000 timesteps.

Additionally, the Cardinality Convergence Curve in \autoref{dst_card} showed that PQL found significantly more solutions and only failed to find one solution. Interestingly, we find that the linear and Chebyshev functions could not retain some solutions found during learning and thus converged with fewer solutions. This is not due to over-training, as performance levels fluctuate before they finally stagnate.

In \autoref{dst_pf}, we noticed that the linear scalarisation function converged with extreme solutions (1,-1) and (18.61,-8.64), which is why it had such a high Hypervolume. The Chebyshev converged with the same solutions and additionally one extra solution (1.96,-4.095) (in some runs, it converges with (13.71, -8.33) or ( 1.62, -2.709) instead). This plot reaffirms the existing literature that linear scalarisation battles to find solutions in non-convex areas of the front. Chebyshev scalarisation handles this substantially better but still does not find all the solutions. Furthermore, we hypothesise that the scalarisation functions are biased to some extent and prefer finding solutions in certain regions of the solution space, so it is unable to retain all solutions during learning. The only solution the Pareto Q-Learning algorithm left undisclosed was (6.78, -7.46).

Due to the linear function only finding extreme solutions yielded the highest Sparsity, as seen in \autoref{dst_sparsity}. On the other hand, Chebyshev tended to find diverse solutions (but failed to retain some of them during learning), so it has much lower Sparsity. In contrast, Pareto Q-Learning discovered most of the solutions, which resulted in significantly low Sparsity scores, meaning it had a dense and uniform coverage of the Pareto front. This correlates to the fact that it has a very low IGD as seen \autoref{dst_igd}. Furthermore, since Chebyshev and linear functions found fewer solutions, which caused considerable gaps in their Pareto front coverage, their IGD scores were notably high.

\subsubsection{Summary of Results}

PQL considerably outperformed the other algorithms. It showed its superiority by almost finding all solutions in significantly fewer timesteps whilst retaining all the solutions found during learning (see \autoref{dst tbl}). On the other hand, this experiment illustrated the linear scalarisation inability to find solutions in non-convex areas of the Pareto front. Chebyshev improved by finding several solutions in the front but did not approximate the entire Pareto front. Notably, with the outer-loop multi-policy evaluation approach of MO Q-Learning, the algorithms allowed it to have 11 distinct weight configurations to approximate the Pareto front, yet it failed to do so. Either the alternate configurations did not converge to any optimal solution, or they found the same optimal solution as the others.
Another key finding was that we found that scalarisation functions often fail to retain all solutions it finds during the learning. The issue is not due to over-training; as evidenced in each learning curve, the performances fluctuate constantly before eventually stagnating. Furthermore, it could reveal some bias with scalarisation functions favouring certain regions of the solution space.
This is a concern, as user preferences may change over time, and we want to have as many optimal solutions on hand as possible. 

\begin{table}[h]
\centering
\begin{tabular}{lcccc}
\toprule
\textbf{Indicator} & \textbf{MOQ-Linear} & \textbf{MOQ-Chebyshev} & \textbf{PQL}& \textbf{         True Pareto front} \\
\midrule
\textbf{Hypervolume $(\uparrow)$} &777.66& 779.281 & \textbf{801.82} &801.842   \\
\textbf{Cardinality $(\uparrow)$} &2  &3  & \textbf{9} &10 \\
\textbf{Sparsity $(\downarrow)$}    &343.79& 259.81  &\textbf{10.017} &8.757  \\
\textbf{IGD $(\downarrow)$}         &4.123&3.611 & \textbf{0.137} &0   \\
\bottomrule

\end{tabular}
\caption{Comparison of algorithms in DST Concave}
\label{dst tbl}
\end{table}

\subsection{Experiment 2: Four-Room}
\label{exp_3}
The previous experiment illustrated that the linear function is less effective in finding solutions in non-convex areas of the Pareto front. The key characteristic of the four-room environment is that it has an unknown Pareto front, which, to some extent, replicates a real-world scenario. This means our optimal trade-off between objectives is not explicitly known, and there are no references or guidelines for comparing and measuring the performance of our algorithms. 

Furthermore, since this objective had 3 objectives, our outer-loop multi-policy approach of the MO Q-Learning function required 64 distinct weight configurations. Refer to Section \ref{MO Q experimental design} for more details. PQL could not be employed in this environment because it struggles to scale to the number of state-action pairs that FourRoom has.

\begin{figure}[h]
    \centering
    \begin{subfigure}[b]{0.48\textwidth}
        \centering
        \includegraphics[width=\linewidth]{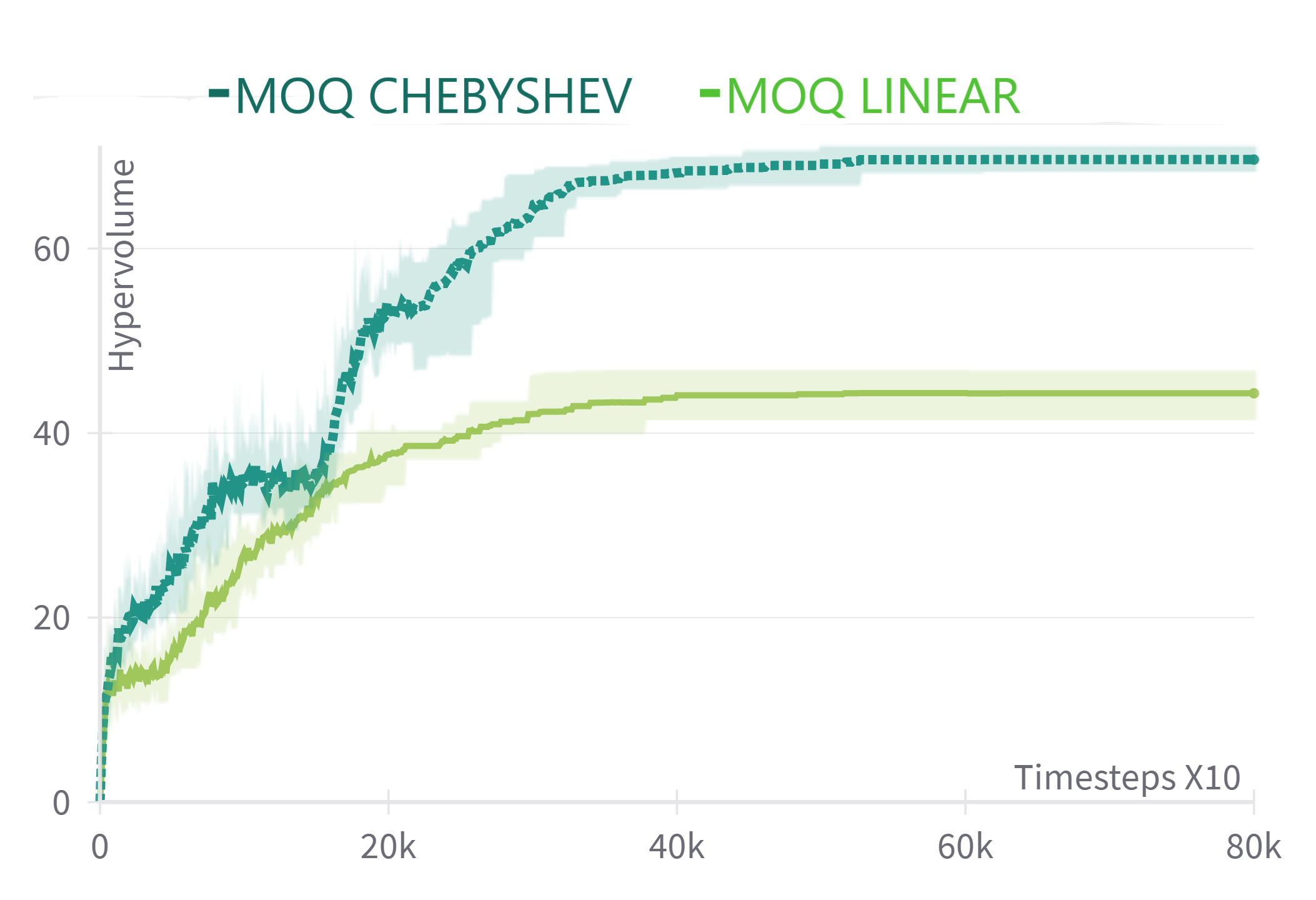}
        \caption{Hypervolume}
        \label{four_room_hypervolume}
    \end{subfigure}
    \hfill
    \begin{subfigure}[b]{0.48\textwidth}
        \centering
        \includegraphics[width=\linewidth]{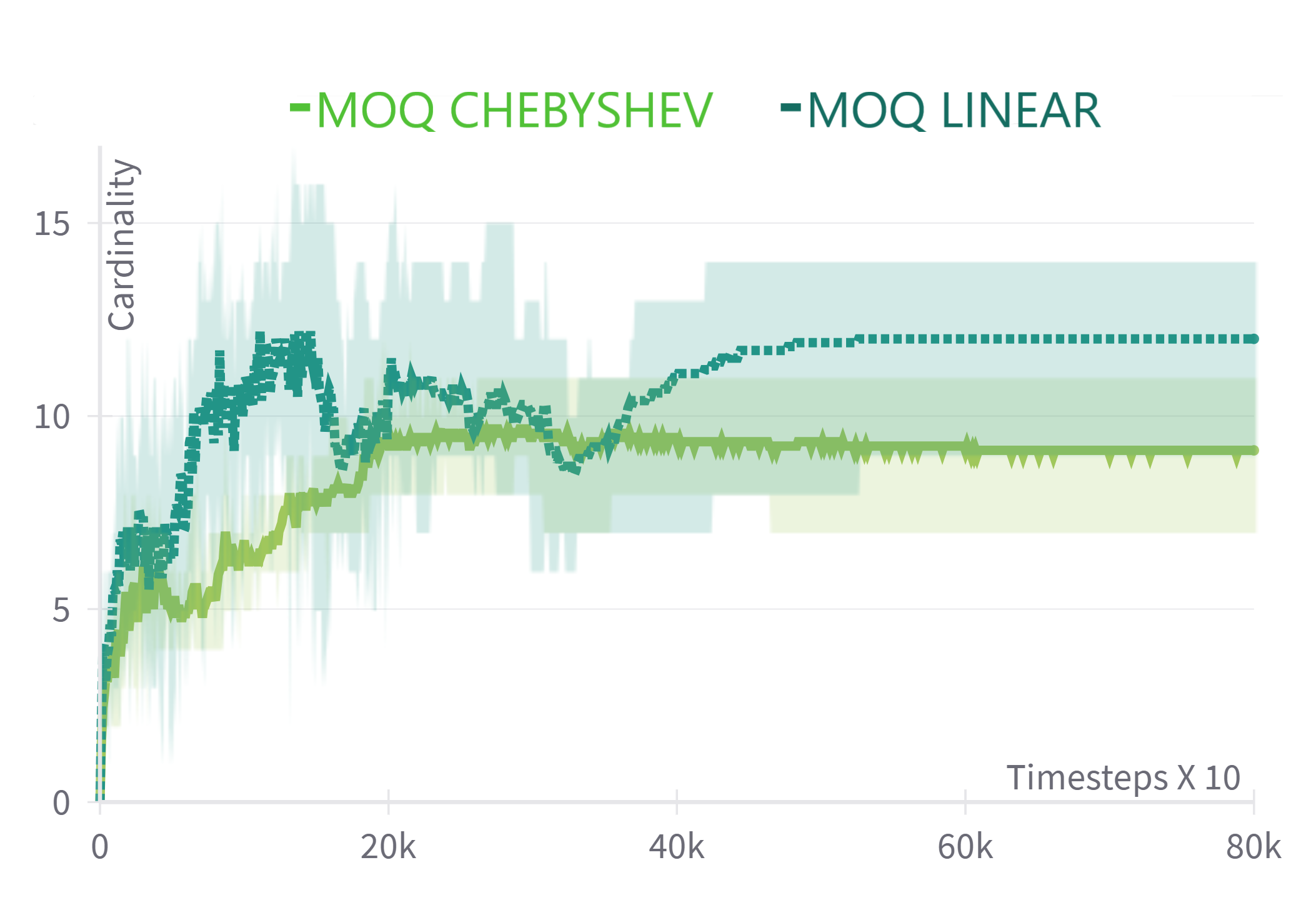}
        \caption{Cardinality}
        \label{four_room_card}
    \end{subfigure}
    \begin{subfigure}[b]{0.48\textwidth}
        \centering
        \includegraphics[width=1\linewidth]{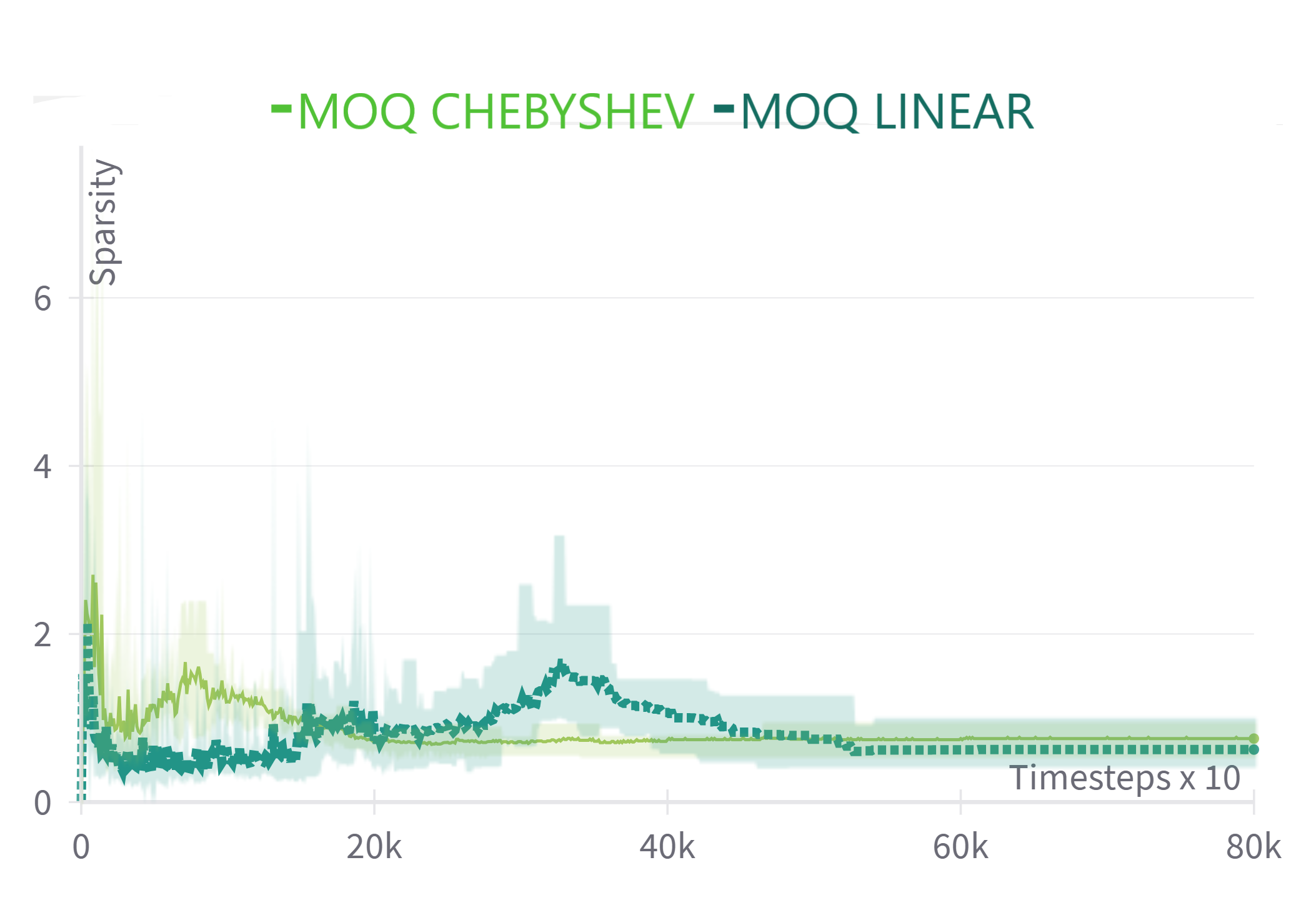 }
        \caption{Sparsity}
    \label{four_room_sparsity}
    \end{subfigure}
    \hfill
    \begin{subfigure}[b]{0.48\textwidth}
        \centering
         \includegraphics[width=1\linewidth]{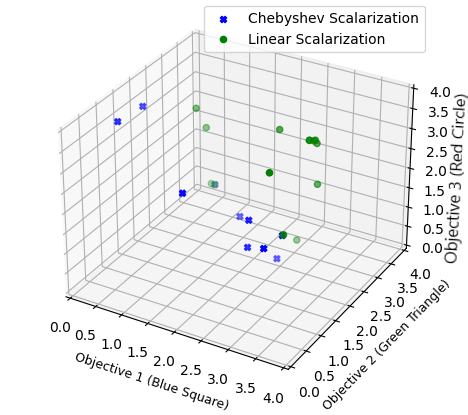}
        \caption{Approximated Pareto Fronts}
    \label{four_room_PF}
    \end{subfigure}
    \caption{Comparison of Algorithms in the Four-Room Environment}
    \label{four_room_comparison}
\end{figure}

\autoref{four_room_hypervolume} shows that the linear scalarisation significantly outperformed the Chebyshev scalarisation, converging with an almost double hypervolume. We also observe that the Chebyshev function converges in much fewer timesteps when compared to the linear function.
We also note that the linear scalarisation uncovered more solutions in \autoref{four_room_card}. Furthermore, the Chebyshev scalarisation stopped discovering new solutions after 400,000 timesteps, while the linear scalarisation continued to find solutions past 600,000. A notable finding was that even though linear scalarisation failed to retain solutions along the way, it found newer solutions, which is why its Hypervolume always increased. In the previous experiment, we observed the same issue, where the scalarised algorithms often identified solutions early during learning but failed to preserve them as training advanced. This is further evidenced in \autoref{four_room_sparsity} where the same Cardinality, resulted in different Sparsity values at different timesteps for each algorithm. This could indicate that the scalarisation functions prefer certain regions in the solution space,

In \autoref{four_room_PF}, we noticed that the solutions uncovered by the Chebyshev function were more diverse and spread out, while the solution set of the linear function was clustered in certain regions. This alluded to the fact that the linear function had lower Sparsity values. Furthermore, the linear function tended to find solutions that aimed at maximising the collection of red circles from the environment. In contrast, Chebyshev found solutions that primarily concentrated on balancing green triangles and blue squares from the environment, although it did have a few solutions that maximised the collection of red circles. Overall, linear scalarisation uncovered solutions with high rewards for all objectives but focused mainly on solutions that revolved around collecting Red Circles. The Chebyshev function finds slightly more diverse solutions but struggles to find solutions that simultaneously yield high rewards for all objectives. It could be due to Chebyshev's prioritising finding solutions that balance all objectives, while linear aims to find solutions that maximise a specific objective, which is why it is focused on maximising the collection of red circles. This also correlates with the point we made in the previous paragraph, which is that the scalarisation function prefers certain regions of the solution space.

\subsubsection{Summary of Results}

Interestingly, while the Chebyshev scalarisation function found more solutions in Experiment 1, in this experiment, the linear scalarisation function was more effective in finding solutions (see Table~\ref{fourroomtable}). This could be due to the fact that the shape of the Pareto front in experiment 1 was entirely non-concave. However, in this environment, while the Pareto front is unknown, solutions don't only lie in non-convex areas. This means that although Chebyshev can find solutions in non-convex points, it does not mean that it can uncover all possible solutions.

Furthermore, this experiment revealed the unreliability of using scalarisation functions, as they converged with entirely different solution sets. This suggests that they may prefer solutions in some areas of the search space. Therefore, it is difficult to determine all the possible solutions with both scalarisation functions converging with entirely different solution sets, which could potentially indicate many more solutions existing in the environment,

Additionally, we employed 64-weight configurations to allow the algorithms to approximate the Pareto front. However, of the 64 weight configurations, the linear scalarisation and Chebyshev scalarisation only found 14 solutions and 11 solutions, respectively, meaning specific configurations converged to identical solutions or could not find any solution. This issue is because we are wasting time and computational resources training different configurations.

An extension of this experiment involves using an inner-loop multi-policy algorithm that is not tabular-based and comparing the solution set found.

\begin{table}[H]
\centering
\begin{tabular}{lcccc}
\toprule
\textbf{Indicator} & \textbf{MOQ-Linear} & \textbf{MOQ-Chebyshev} \\
\midrule
\textbf{Hypervolume$(\uparrow)$} &\textbf{71.38}& 49  \\
\textbf{Cardinality $(\uparrow)$} &\textbf{12}  &9 \\
\textbf{Sparsity $(\downarrow)$}   &\textbf{0.625}& 0.756   \\

\bottomrule

\end{tabular}
\caption{Comparison of scalarisation methods.}
\label{fourroomtable}
\end{table}

\section{Conclusion}
\label{conclusion}
Chebyshev scalarisation overcame certain limitations associated with linear scalarisation, particularly discovering solutions in non-convex regions of the Pareto front. However, the spread and coverage of the solutions found by each scalarisation function highly depend on the environment's characteristics. In Experiment 1, the Chebyshev scalarisation function performed better, while in Experiment 2, the linear function was more effective. 

Furthermore, scalarisation functions fail to preserve solutions during learning and often overlook many optimal solutions as they tend to favour finding solutions in certain regions of the search space. In experiment 2, both functions converged with entirely different sets, limiting their reliability and stability for real-world applications.

Additionally, finding the appropriate weight configurations to sample the entire Pareto front accurately is challenging, with many distinct configurations converging with the solutions already discovered, wasting computational resources. Therefore, finding the appropriate weight configurations to sample the entire Pareto front is complex, making outer-loop multi-policy method approaches less effective. Additionally, as the number of objectives in settings expands, it becomes increasingly more complicated to approximate the Pareto front, and the procedure adopted in this paper to calculate the possible weight configurations would be unsustainable and computationally infeasible. It would be more computationally viable to have an algorithm that could learn all the solutions in a single run which is what inner-loop multi-policy algorithms aim to achieve. All these points mentioned above highlight their limitations for intelligent decision-making.

In contrast, the inner-loop multi-policy method demonstrated its efficacy by converging to almost the entire Pareto optimal set in experiment 1. Based on the results, we encourage the change from outer-loop policy methods to inner-loop multi-policy to develop more sophisticated and intelligent systems that can adapt to real-world complexities\citealp[]{felten2024multi,roijers2016multi,lu2023multi}.

\section{Future Work}
\label{futurework}

This comparative study focused on primarily assessing scalarisation techniques employed in MORL. Although we demonstrated the limitations of this approach, using additional environments with various action and observation spaces, such as continuous and pixel spaces, would further expand the results and insights obtained in this study. Furthermore, we wish to assess the limitations of newer inner-loop multi-policy MORL algorithms such as MORL/D \cite{felten2024multi}, Optimistic Linear Support (OLS) \cite{roijers2016multi} and CAPQL \cite{lu2023multi} which will allow for more intelligent and robust systems.

\bibliographystyle{plain}

\appendix
\section*{Appendix}
\label{appendix}

A parameter search was performed to find the optimal parameters for each algorithm in the environment. The Weights \& Biases framework was used to automate the process, which is referred to as a sweep.
\label{append}
\section{Sweep 1: Deep-Sea Treasure Concave}
\label{sweep dst}

\begin{figure}
    \centering
    \includegraphics[width=0.8\linewidth]{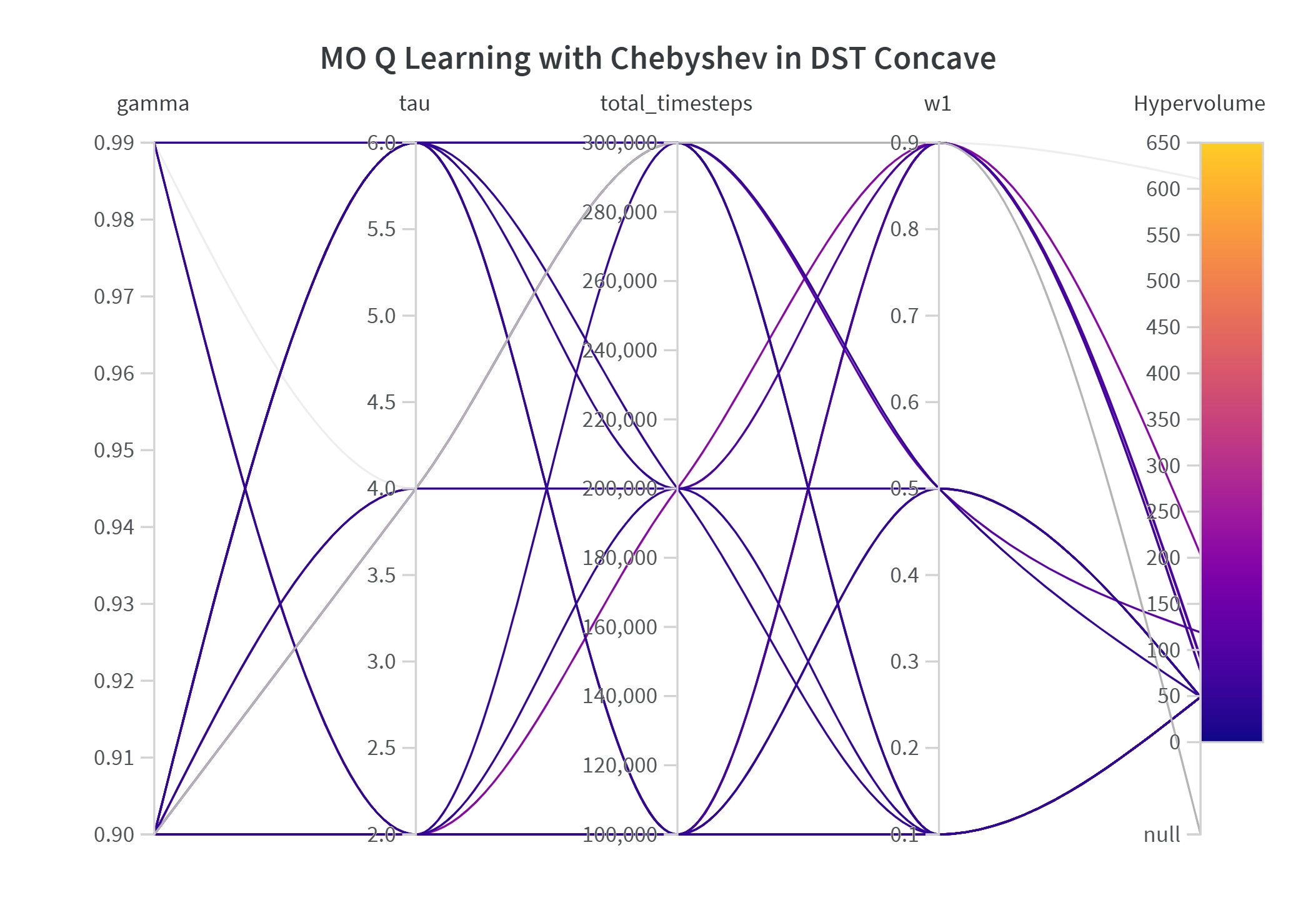}
        \caption{Parameter Sweep for MO Q-Learning with Chebyshev Scalarisation in DST Concave}
    \label{dst sweep cheb moq}
\end{figure}

In this environment, the interest is finding deeper than immediate solutions. The initial parameter sweep for Chebyshev scalarisation consisted of tau values of 1,2,4,6, total timesteps set to 100000, 200000, and 300000, and discount factors $\gamma$ of 0.8, 0.9 and 0.99. With a gamma rate of 0.8, the algorithms performed poorly and could not find optimal solutions. The $\gamma$ of 0.8 was removed from the sweep to focus on the other parameters.

In \autoref{dst sweep cheb moq}, we observed with fewer timesteps, i.e. 100 000 and 200 000; the algorithm often converged to treasures in shallower regions. With $\gamma$ rates of 0.9 and 0.99, we were getting similar reward returns. However, the Hypervolume scores were completely different, with the solutions having a $\gamma$ of 0.99 higher. This is because the shape of the Pareto front is calculated based on the discounted vector reward, which is directly influenced by $\gamma$. While a $\gamma$ of 0.99 yielded higher Hypervolume, the algorithms performed similarly with a $\gamma$ of 0.9. With a $\gamma$ of 0.99, the true Hypervolume was 3208; with 0.9, it was 854. Thus, we chose to use a $\gamma$ of 0.9. Furthermore, the $\tau$ parameter was insensitive to the solutions found.

\begin{figure}
    \centering
    \includegraphics[width=0.8\linewidth]{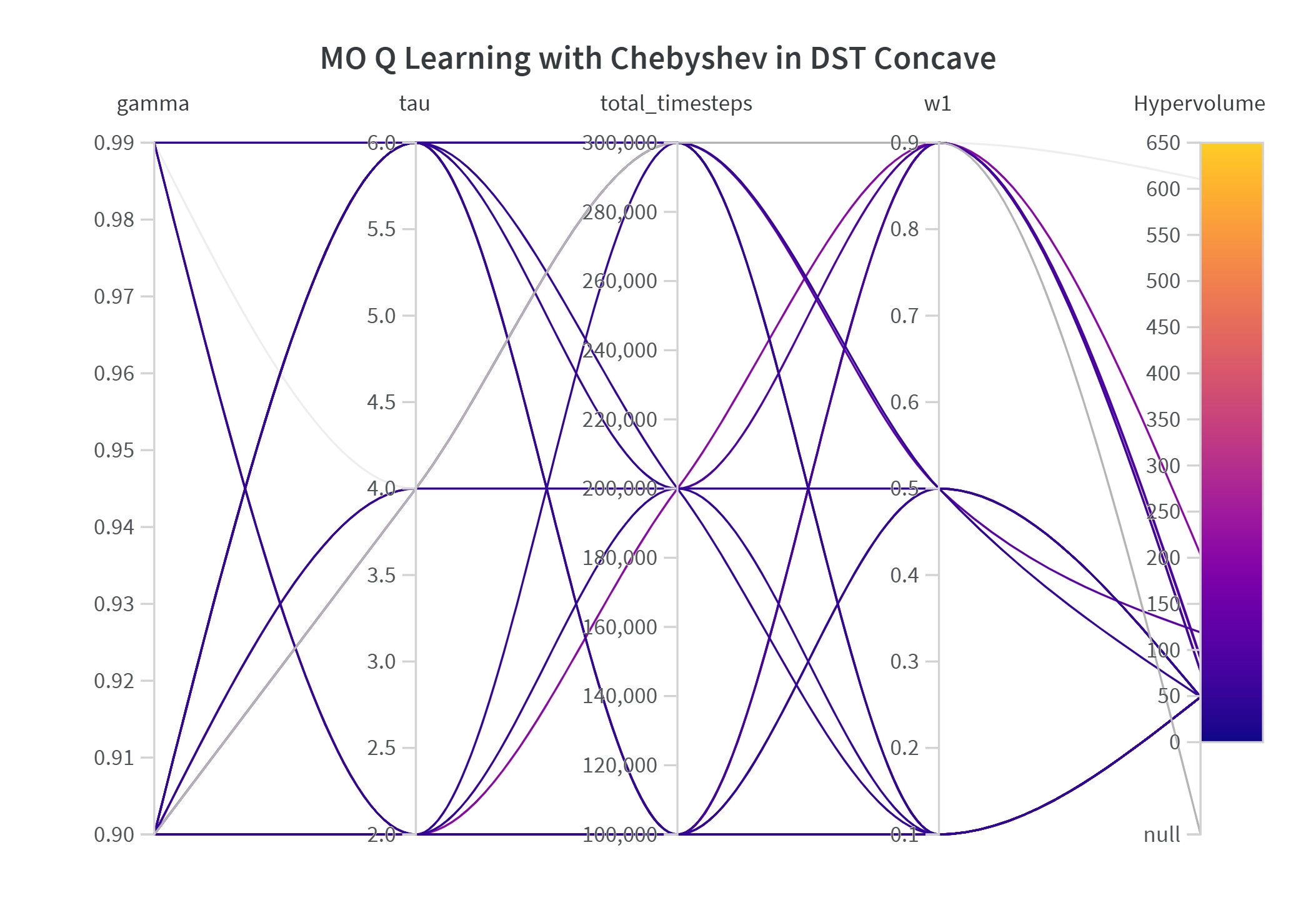}
        \caption{Parameter Sweep for MO Q-Learning with Linear Scalarisation in DST Concave}
    \label{linear sweep}
\end{figure}
With the linear scalarisation parameter sweep in \autoref{linear sweep}, the $\gamma$ was set to 0.9 and 0.99, with timesteps of 100000, 200000 and 300 000. Again, it was found that utilising more timesteps yielded deeper treasure solutions with higher Hypervolume.. With timesteps of 300,000, the algorithm found lucrative treasures with better reward returns and higher Hypervolume. Therefore, a $\gamma$ of 0.9 and 400,000 as the total timesteps was used. The  $\tau$ parameter of the Chebyshev function was set to 4.

\subsubsection{Sweep 2: Four-Room}
\label{four room sweep}
Since the Pareto front is unknown in this environment, we ran multiple sweeps to consolidate the chosen parameters.

In the initial parameter sweep for Chebyshev Scalarisation, the $\tau$  was 1,2,4,6, $\gamma$ to 0.9 and 0.99 with timesteps set to 500 000. 
\begin{figure}
    \centering
    \includegraphics[width=0.8\linewidth]{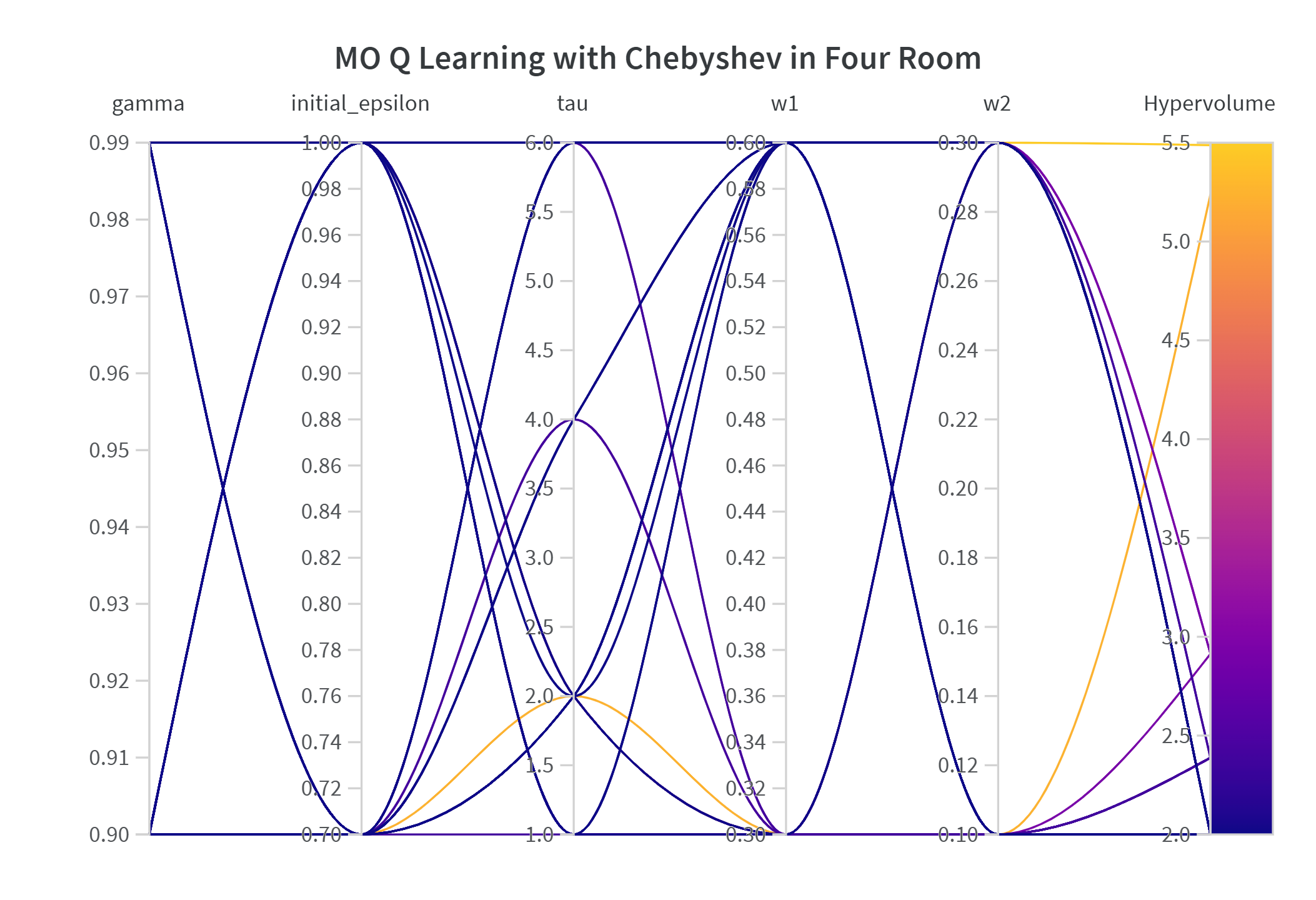}
        \caption{Parameter Sweep for MO Q-Learning with Chebyshev Scalarisation in Four-Room}
    \label{sweep cheby four room}
\end{figure}
In \autoref{sweep cheby four room}, we observed that the algorithm became more future-oriented with a $\gamma$ of 0.99 and found higher-yielding reward returns along with a higher Hypervolume. Once again, it was observed that the tau parameter was insensitive to the solutions found.

With the linear scalarisation parameter sweep, the $\gamma$ was configured to 0.9 and 0.99, with timesteps set to 500 000. 
\begin{figure}
    \centering
    \includegraphics[width=0.8\linewidth]{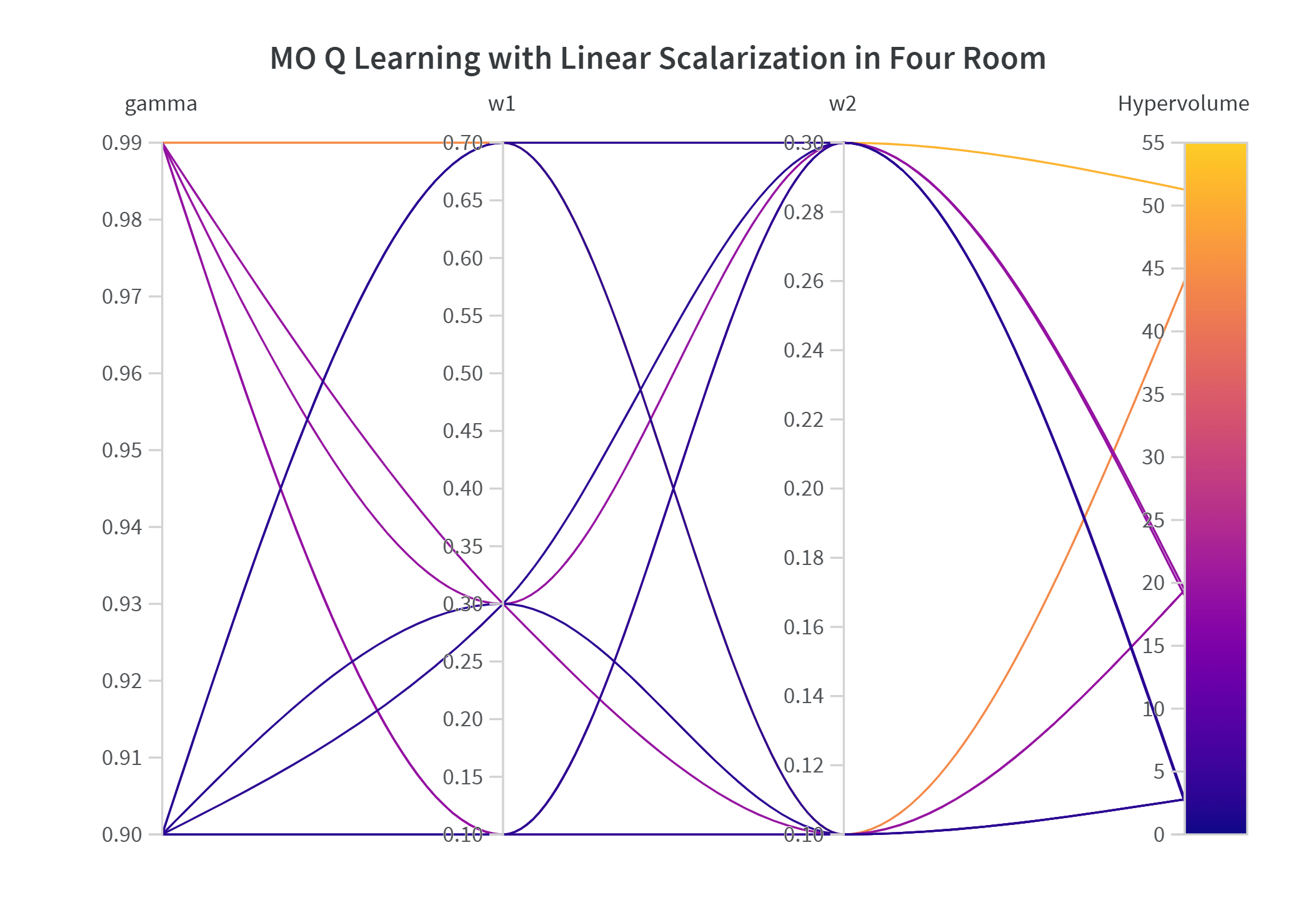}
        \caption{Parameter Sweep for MO Q-Learning with Linear Scalarisation in Four-Room}
    \label{sweep linear four room}
\end{figure}
Again, in \autoref{sweep linear four room}, it was found that a higher $\gamma$ rate yielded higher reward returns.

After observing the above, we investigated whether reducing the exploration would yield exact optimal solutions as a full $\epsilon$ decaying exploration approach. Another parameter sweep was performed for the Chebyshev function and decayed our $\epsilon$ over different rates of 0.1, 0.2,0.5 and 1 with 500 000 timesteps.

\begin{figure}
    \centering
    \includegraphics[width=0.8\linewidth]{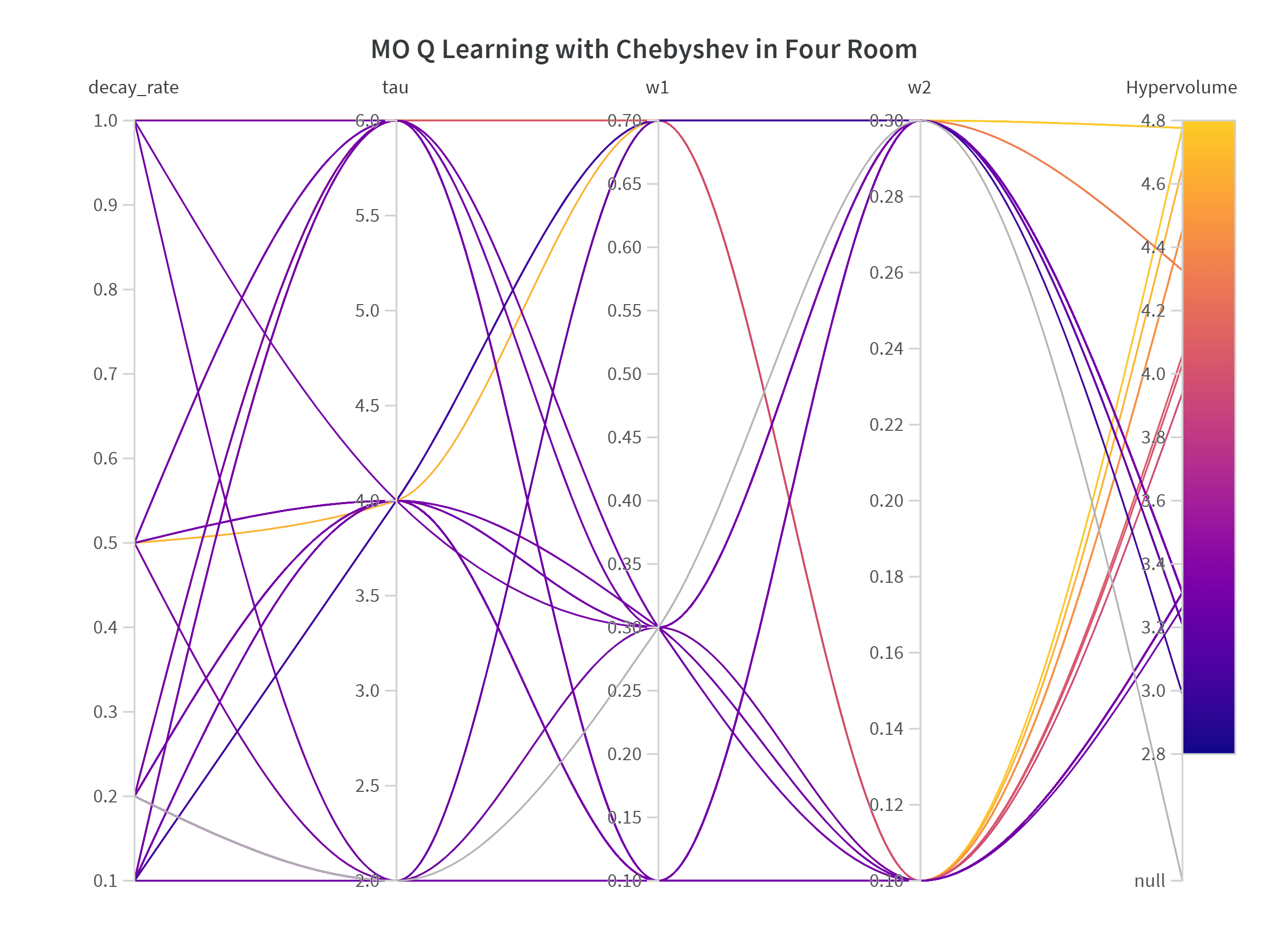}
        \caption{Parameter Sweep for MO Q-Learning with Chebyshev Scalarisation in Four-Room}
    \label{sweep cheby2 four room}
\end{figure}
In \autoref{sweep cheby2 four room}, we noticed that the more the algorithm explored, the higher the reward returns and Hypervolume. Thus, a discount factor $\gamma$ of 0.99 was used in the Four-Room environment. A full epsilon decaying approach was adopted, which started from 1 and gradually decayed to 0.1 over 800 00 times (decay rate was 1),  while the $\tau$ parameter for the Chebyshev mechanism was set to 6.

\end{document}